\documentclass[journal]{IEEEtran}

\pdfoutput=1

\usepackage{cite}
\usepackage[pdftex]{graphicx}
\DeclareGraphicsExtensions{.pdf,.jpeg,.png}
\usepackage{amsmath}
\usepackage[noend]{algpseudocode} %loads algorithmicx
\usepackage{array}
  \usepackage[caption=false,font=footnotesize]{subfig}
\usepackage{algorithm} %bare_jrnl.tex (in IEEEtran) says do not use algorithm environment?
\usepackage{amssymb}
\usepackage{amsthm}
\usepackage{pgfplots}

\pgfplotsset{compat=1.13}

\algnewcommand\algorithmicswitch{\textbf{switch}}
\algnewcommand\algorithmiccase{\textbf{case}}
\algnewcommand\algorithmicassert{\texttt{assert}}
\algnewcommand\Assert[1]{\State \algorithmicassert(#1)}%
\algdef{SE}[SWITCH]{Switch}{EndSwitch}[1]{\algorithmicswitch\ #1\ \algorithmicdo}{\algorithmicend\ \algorithmicswitch}%
\algdef{SE}[CASE]{Case}{EndCase}[1]{\algorithmiccase\ #1}{\algorithmicend\ \algorithmiccase}%
\algtext*{EndSwitch}%
\algtext*{EndCase}%

\makeatletter
\newcommand{\StatexIndent}[1][3]{%
  \setlength\@tempdima{\algorithmicindent}%
  \Statex\hskip\dimexpr#1\@tempdima\relax}
\algdef{S}[WHILE]{WhileNoDo}[1]{\algorithmicwhile\ #1}%
\makeatother

\newcommand\copyrighttext{%
	\small \textcopyright 2019 IEEE.  Personal use of this material is permitted.  Permission from IEEE must be obtained for all other uses, in any current or future media, including reprinting/republishing this material for advertising or promotional purposes, creating new collective works, for resale or redistribution to servers or lists, or reuse of any copyrighted component of this work in other works.}
\newcommand\copyrightnotice{%
	\begin{tikzpicture}[remember picture,overlay]
	\node[anchor=north,yshift=-10pt] at (current page.north) {{\parbox{\dimexpr\textwidth-\fboxsep-\fboxrule\relax}{\copyrighttext}}};
	\end{tikzpicture}%
}

\begin{document}

\title{Persistent Multi-UAV Surveillance with Data Latency Constraints}

\author{J\"urgen~Scherer,
        Bernhard~Rinner$^{1}$%
\thanks{$^{1}$Both authors are with the Institute of Networked and Embedded Systems, Alpen-Adria-Universit\"at Klagenfurt, Austria,
{\tt\footnotesize \{juergen.scherer, bernhard.rinner\}@aau.at} }
}

\maketitle
\copyrightnotice

%\thispagestyle{empty}
%\pagestyle{empty}

%%%%%%%%%%%%%%%%%%%%%%%%%%%%%%%%%%%%%%%%%%%%%%%%%%%%%%%%%%%%%%%%%%%%%%%%%%%%%%%%
\begin{abstract}
We discuss surveillance with multiple unmanned aerial vehicles (UAV) that minimize idleness (the time between consecutive visits of sensing locations) and constrain latency (the time between capturing data at a sensing location and its arrival at the base station). This is important in persistent surveillance scenarios where sensing locations should not only be visited periodically, but the captured data also should reach the base station in due time even if the area is larger than the communication range. Our approach employs the concept of minimum-latency paths (MLP) to guarantee that the data reaches the base station within a predefined latency bound. To reach the bound, multiple UAVs cooperatively transport the data in a store-and-forward fashion. Additionally, MLPs specify a lower bound for any latency minimization problem where multiple mobile agents transport data in a store-and-forward fashion. We introduce three variations of a heuristic employing MLPs and compare their performance in a simulation study. The results show that extensions of the simplest of our approaches, where data is transported after each visit of a sensing location, show improved performance and the tradeoff between latency and idleness.
\end{abstract}

%\category{I.2.9}{Artificial Intelligence}{Robotics}[autonomous vehicles]
%\terms{Design, Experimentation}
%\keywords{Unmanned aerial vehicles; UAV; surveillance; aerial communication; scheduling; path planning}


\IEEEpeerreviewmaketitle

\section{Introduction}
\label{sec:introduction}

\IEEEPARstart{A}{dvances} in the field of aerial robotics have lead to a great interest in the use of unmanned aerial vehicles (UAVs) for various civilian applications. Potential applications for multi-UAV systems include surveillance \cite{Meng2015}, disaster response \cite{Erdelj2017}, \cite{Scherer2015}, \cite{Khan2018}, wildfire monitoring \cite{Ghamry2016} and environmental monitoring \cite{Rossi2016}.

In this work we consider a path planning problem for multiple UAVs (or other types of mobile robots) that are visiting points of interest (which we denote as sensing locations) periodically over a longer period of time, which is typically required in disaster response scenarios. Since the area is potentially large, existing wireless communication technology used on aerial vehicles is not able to establish a fully connected network all the time due to range limitations while the UAVs move to the sensing locations. Direct communication to the base station (BS) could also be prohibited while flying close to the ground or within a building. In these scenarios it is not only important that sensing locations get visited periodically, but also that the data captured by the UAV arrives at the base station in due time. This allows human mission operators to quickly assess a situation or that the collected data can be promptly processed for further analysis.

\begin{figure}
	\centering
	\includegraphics[scale=0.25]{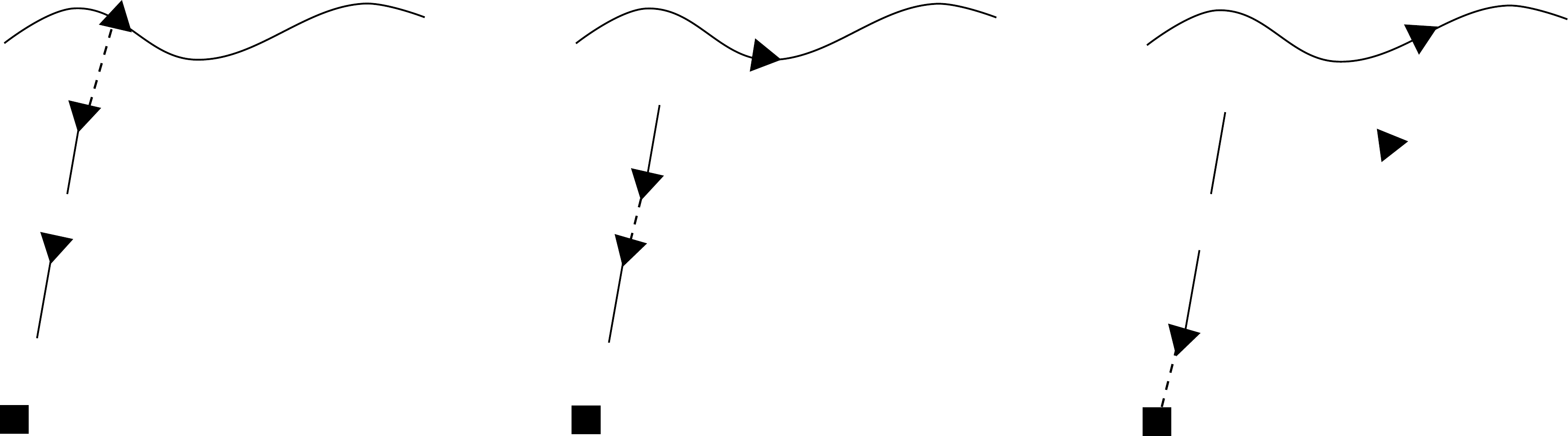}
	\caption{Three points in time (left to right) with UAVs (triangles) and the BS (square). The top UAV is visiting sensing locations along a path (waved line) and the other UAVs collaboratively transport the data to the BS on minimum-latency paths (straight solid line). Dashed lines indicate that UAVs exchange data.}
	\label{fig:scenario}
\end{figure}

We present a novel collaborative data delivery approach where UAVs transport data in a store-and-forward fashion. This is different from our previous work \cite{Scherer2016}, \cite{Scherer2017} where permanent connectivity to the BS was considered. The paths for the UAVs are planned such that two UAVs meet at points that are within their communication range. The data is sent from one UAV to the other, which travels along a path to meet other UAVs. By following so called \textit{minimum-latency paths} (MLP) the data travels towards the BS with a guaranteed latency while minimizing the time required for visiting all sensing locations at least once (see Figure~\ref{fig:scenario} for an illustration). Repeating the generated path leads to a persistent surveillance path that tries to minimizes the idleness over all sensing locations. Idleness is defined as the time between two consecutive visits at a sensing location, and latency is defined as the time between the capture of the data at a sensing location and its arrival at the BS. We call this problem \textit{minimum idleness with latency constraints} (MILC).

The rationale behind cooperative data transportation is that UAVs do not have to travel to the BS individually to deliver the data but can spent more time on visiting sensing locations while the previously captured data travels to the BS in a coordinated way over multiple UAVs. This eliminates large detours if the communication range is large compared to the travel speed or if no fly zones block movement but not communication. Surveillance scenarios where data gathering tasks are defined by mission operators on demand are other motivating examples for cooperative data transport along MLPs. A newly arising task is approached by a UAV, which gathers the data and available UAVs transport the data as fast as possible to the base station. Here, the rationale is that not all UAVs are employed permanently during the whole mission and spare UAVs can assist in transporting the data.

The remainder of the paper is organized as follows: In Section~\ref{sec:relatedwork} we review the existing literature. In Section~\ref{sec:problem} we introduce MILC and some notation. In Section~\ref{sec:algorithm} we introduce MLPs and describe three heuristics for MILC. Section~\ref{sec:eval} describes the simulation results and Section~\ref{sec:conclusion} concludes the article.

\section{Related Work}
\label{sec:relatedwork}%\cite{Fargeas2013}
While minimizing idleness in persistent surveillance and patrolling applications is a common optimization goal \cite{Nigam2012}, \cite{Fargeas2013}, \cite{Portugal2014}, minimizing latency has received much less attention in literature. Banfi et al. \cite{Banfi2015} present a MILP (mixed integer linear program) formulation and heuristics for the problem of finding a patrolling path for each UAV with the goal to minimize the latency. Each UAV follows a path containing sensing locations and intermediate detours to communication sites where the data can be transmitted to the BS. A MILP formulation and a heuristic for a similar problem with task revisit constraints are presented in \cite{Manyam2017}. Acevedo et al. \cite{Acevedo2013_ICUAS} investigates in patrolling considering the propagation of information among the UAVs. A decentralized algorithm maintains a grid shaped partition of the area where each UAV is traveling along a circular path within its subarea. UAVs exchange data on the border of their subareas with each UAV of the neighboring subareas, which minimizes the propagation time of information in this grid shaped partition.

In contrast to related work, we consider a collaborative data transport by multiple UAVs to a single BS. Other work focuses on recurrent connectivity without explicitly minimizing data latency \cite{Pasqualetti2012b}, \cite{Flushing2013}, \cite{Kantaros2019}. Recurrent connectivity of the full robot network, e.g. with the aim for planning and coordination of the next tasks, has been considered in \cite{Hollinger2012}, \cite{Banfi2018}.

\section{Problem description}
\label{sec:problem}

In this section we formally define the MILC problem. The set of UAVs is denoted as $R=\{r_1, \ldots, r_n\}$, with $|R|=n$. The problem is modeled with help of a weighted movement graph $G_M=(V, E_M, W^M)$ where the UAVs can move from one vertex $v$ to another $w$ within time $W^M_{vw}$ if there is an edge $(v, w) \in E_M$. Vertex $v_0\in V$ identifies the BS. The set of vertices $V_S \subseteq V$ are sensing locations which have to be visited by at least one UAV. The set $V_C \subseteq V_S$ contains sensing locations in communication range of the BS. The communication graph $G_C=(V, E_C, W^C)$ models the communication connectivity between vertices. If there is an egde $(v, w) \in E_C$ then two UAVs or the BS and a UAV can communicate if one is at $v$ and the other is at $w$ at the same time. The edge weights $W^C$ describe the duration of the data transmission for every edge. The latency for a certain sensing location is defined as the time between the collection of the data at a sensing location and the arrival of the data at the BS. To collect data at $v\in V_S$ a UAV must be at $v$ but data is not necessarly collected each time a UAV is at $v$. Data collection is scheduled such that the latency $L^{constr}$ can be ensured. The objective is to minimize the time all $v \in V_S$ have been visited while maintaining the latency constraint $L^{constr}$.

Determining the optimal tour for minimal idleness on a graph is related to the traveling salesperson problem (TSP) \cite{Garey1979} and the k-TSP \cite{Frederickson1976}, which are both NP-complete. Since MILC with one UAV and a sufficiently large latency bound (e.g. sum of all edge weights) is equivalent to a TSP, MILC is NP-hard too. In the next section we describe heuristic algorithms for solving this problem efficiently.

\section{Algorithm description}
\label{sec:algorithm}

\subsection{Minimum-latency path}
\label{subsec:minlatencypath}

The problem of transporting the data as fast as possible from a source $s \in V$ to a destination $d \in V$ with a given number of UAVs can be modeled as a shortest path problem with time windows (SPPTW) in a graph $G=(V,A)$. We call this path minimum-latency path (MLP). SPPTW is the problem of finding a shortest path (cycles are allowed) in a graph with a traversal cost $W_{vw}$ and a traversal duration $T_{vw}$ associated with every edge $(v,w) \in A$. The sum of edge traversal times along the path is constrained to be in a time window $[L_v, U_v]$ at a vertex $v \in V$. The problem without cycles can be formulated as follows:

\begin{equation}
	\min \sum_{(v,w)\in A}{W_{vw} x_{vw}}
\end{equation}
\begin{equation}
	\label{eq:spptw_flow}
	\sum_{w \in V}{x_{vw}} - \sum_{w \in V}{x_{wv}} = 
	\begin{cases} 
		+1 & v = s \\
		\hphantom{+} 0 & i \in V \setminus \{s,d\} \\
		-1 & v = d
	\end{cases}
	\qquad \forall v \in V
\end{equation}
\begin{equation}
	x_{vw} \geq 0 \qquad \forall (v,w) \in A
\end{equation}
\begin{equation}
	\label{eq:spptw_t}
	x_{vw} \cdot (t_v + T_{vw} - t_w) \leq 0 \qquad \forall (v,w) \in A
\end{equation}
\begin{equation}
	\label{eq:spptw_tw}
	L_v \leq t_v \leq U_v \qquad \forall v \in V
\end{equation}

This nonlinear formulation contains two types of variables: the flow variables $x_{vw}, (v,w) \in A$, and time variables $t_v, v \in V$. The flow constraints (\ref{eq:spptw_flow}) ensure a path from $s$ to $d$, and constraints (\ref{eq:spptw_t}) and (\ref{eq:spptw_tw}) enforce a visit at $i$ within a time window. Note that if an arrival happens before $L_v$, a waiting time is introduced. 

The problem is NP-hard in general, but there exist dynamic programming algorithms for the problem instance with integer $T_{ij}$ \cite{Desrosiers1995}. The dynamic programming approach converts the problem to a shortest path problem in a graph with at most $\sum_{v \in V}{\left( U_v - L_v + 1 \right) }$ vertices. In our case, $W_{ij}$ is the time it takes to traverse the edge $(v,w) \in E_M$, and $T_{vw}$ is the number of UAVs necessary to traverse the edge (1 for every edge $(v,w) \in E_C$), the lower limit is $L_v=0$ and the upper limit is $U_v=|R|-1$ for all vertices.

\begin{figure}[t]
	\centering
	\begin{tabular}{cc}
		\subfloat[]{
			\label{fig:example_spptw_graph}
			\includegraphics[scale=0.38]{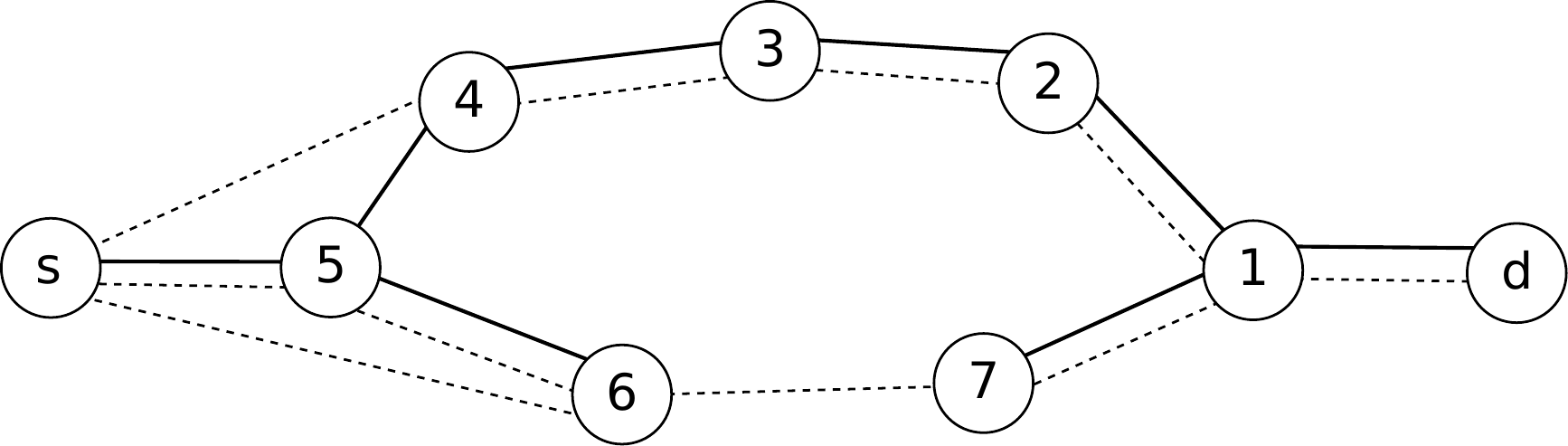}
		}
		\\
		\subfloat[]{
			\label{fig:example_spptw_dp}
			\includegraphics[scale=0.38]{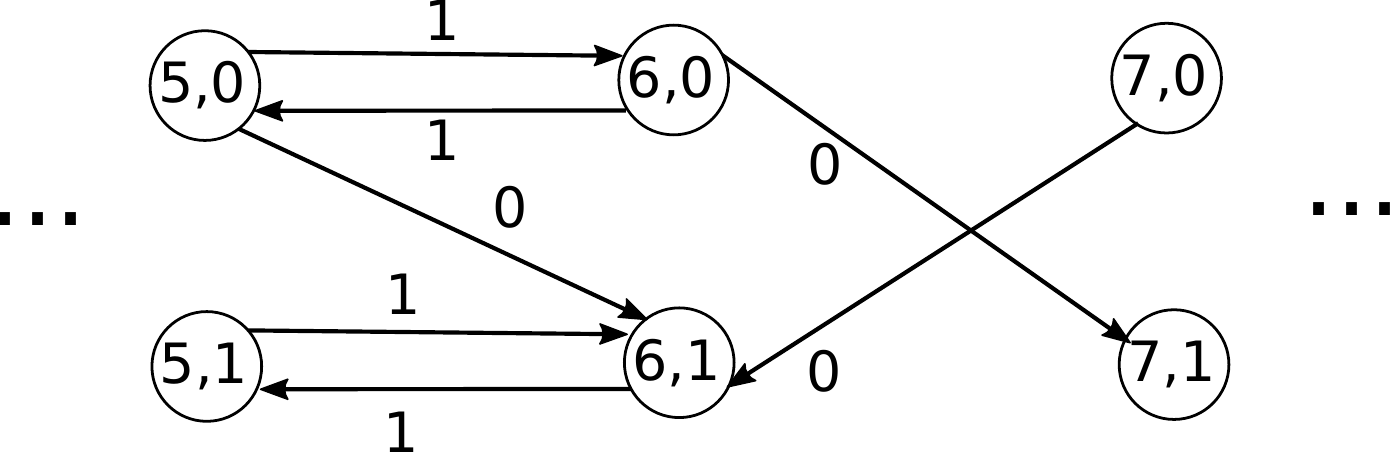}
		}
	\end{tabular}
	\caption{(a) Graph with edges from $E_M$ (solid lines, all $W^M$ are 1) and edges from $E_C$ (dashed lines, all $W^C$ are 0). (b) Converted graph from (a) for the dynamic programming approach of SPPTW with $|R|=2$. Only the vertices and edges corresponding to the vertices $5,6,7$ are shown.}
	\label{fig:example_spptw}
\end{figure}

Figure~\ref{fig:example_spptw_graph} depicts an example graph. If there are two types of edges (from $E_M$ and $E_C$) between two vertices $v$ and $w$, there are two possible ways to transport the data from $v$ to $w$. Either one UAV moves from $v$ to $w$, which takes $W^M_{vw}$ time, or one UAV is placed at $v$ and the other at $w$ and the data is transmitted in time $W^C_{vw}$. In the latter case the edge ``consumes'' one UAV. In our example, with $W^M_{vw}=1$ and $W^C_{vw}=0$ for all $(v,w)\in A$, there are different optimal paths in terms of the latency depending on the number of available UAVs. If $|R|=1$, the MLP from $s$ to $d$ is $\left( s, 5, 4, 3, 2, 1 (stop) \right)$ with a latency of $5$. This notation means that the UAV moves from $s$ to $1$ along the solid edges and stops at vertex $1$ to transmit the data to the destination $d$. If $|R|=2$, the MLP is $\left( s, 5, 6 (stop), 7, 1 (stop)\right)$ or $\left( s (stop), 4, 3, 2, 1 (stop) \right)$ with a latency of $3$. In the first case the first UAV moves from $s$ over $5$ to $6$ and stops there to transmit the data to the second UAV waiting at $7$. This UAV finally transports the data to $1$. In the second case the first UAV does not move at all and transfers the data to the second one waiting at $4$. Finally, if $|R|=3$, the optimal path is $\left( s (stop), 6 (stop), 7, 1 (stop) \right)$ with a latency of $1$.

Figure~\ref{fig:example_spptw_dp} shows parts of the converted graph for $|R|=2$. The vertex $(5,0)$ on a valid path from $s$ to $d$ represents the fact that only one UAV has been used so far. From this vertex the path can continue over $(6,0)$, which means that the edge $(5,6) \in E_M$ is used. If the path continues to vertex $(6,1)$, then the edge $(5,6) \in E_C$ is used, which means that the first UAV stops at $5$ and transmits the data to $6$. From $(6,1)$ there is no path to $(7,0)$ or $(7,1)$ because $(6,7) \notin E_M$ and two UAVs have already been used. The edge length in the converted problem is $1$ for a movement on an edge from $E_M$ or $0$ for transmitting the data over an edge from $E_C$. The problem of finding the shortest latency path reduces to finding the shortest path in the graph of the converted problem.

The MLP represents a lower bound for any problem of minimizing the latency where the data is transported in a store-and-forward fashion by multiple mobile agents. Computing an MLP when the number of agents is $|R|$, the number of vertices is $|V|$, and the edges $E_C$ and $E_M$ are given as adjacency matrices has time complexity $O(|V|^2 \cdot |R|)$ for generating the adjacency matrix for the new graph (with $|V| \cdot |R|$ vertices), and $O(|V|^2 \cdot |R|^2)$ for calculating the shortest path (using Dijkstra's algorithm) in this new graph.

\subsection{Heuristics for MILC}
\label{subsec:milc-h}

The basic idea of the heuristics is that for each $v\in V_S$ UAVs move to the initial positions along the MLP from $v$ to the BS $v_0$ and wait, if necessary, for its preceding UAV on the MLP that transmits the data captured at $v$. It then moves to its final position on the MLP to transmit the data to its successor on the MLP. The order at which sensing locations are visited is determined by the TSP tour. The output of the algorithms are mappings from UAVs to its start vertex ($sv_v$), end vertex ($ev_v$), and start time ($st_v$) on a MLP for every vertex $v \in V_S$. The start time determines the time that it has to wait for its preceding UAV on the MLP to transmit the data of $v$.

\begin{algorithm}[t]
\caption{Heuristic for MILC (MILC-H1)}
\label{alg:milc_h1}
%\footnotesize
\begin{algorithmic}[1]
	\Require
		\Statex movement and communication graphs $G_M, G_C$, sensing locations $V_S$, number of UAVs $|R|$, latency bound $L^{constr}$
	\Ensure
		\Statex subtours $t_1, \ldots, t_k$
		\Statex MLP and schedule $(st(v), sv_{v}, ev_{v}, st_{v}) \quad \forall v \in V_S$
	\Statex

	\For{$v \in V_S$}
		\State $r_v \gets \infty$
		%\State /* Can also be done with interval bisection method ($\rightarrow O(\log |R|)$ iterations) */
		\State /* $O(\log |R|)$ iterations with interval bisection: */
		\For{$i = |R| \textbf{ downto } 1$}\label{line:milc_h1_loopminrobot}
			\State $d \gets min\_latency(v, v_0, i, G_M, G_C)$
			\If{$d \leq L^{constr}$} $r_v \gets i$\label{line:milc_h1_minrobot}
			\Else $\:$ break
			\EndIf
		\EndFor
		\If{$r_v > |R|$}\label{line:milc_h1_infeasible} exit ``Problem is infeasible!''
		\EndIf
	\EndFor
	\State $r \gets \max_{v \in V_S \setminus V_C}{\{r_v\}}$\label{line:milc_h1_maxr}
	\State $k \gets \lfloor |R|/r \rfloor$
	\State $T \gets solve\_tsp(V_S, G_M)$
	\State $(t_1, \ldots, t_k) \gets split\_tour(k, T)$\label{line:milc_h1_splittour}
	\For{$i = 1 \textbf{ to } k$}\label{line:milc_h1_fortours}
		\State $R_i \gets \{(i-1)\cdot r+1, \ldots, i \cdot r\}$
		\State $v' \gets v_0$
		%\For{$j = 1 \textbf{ to } |t_i|$}
		\For{each $v$ on path $t_i$}	\label{line:milc_h1_forpath}
			%\State $v' \gets v$
			%\State $v \gets t_i(j)$
			\State $(sv_{v}, ev_{v}, st_{v}, et_{v})$
			\StatexIndent[3] $\gets min\_latency\_path(v, v_0, r_v, G_M, G_C)$\label{line:milc_h1_minlatencypath}
			\For{$l = 1 \textbf{ to } r_v$, $\forall m \in R_i$}\label{line:milc_h1_calcm}
				%\For{$m \in R_i$}
					\State $A_{lm} \gets st(v') + et_{v'}(m) + $
					\StatexIndent[5] $dist_{G_M}(ev_{v'}(m), sv_{v}(l))$
				%\EndFor
			\EndFor
			\State $M \gets minmax\_matching(A)$\label{line:milc_h1_minmax}
			\For{$m \in R_i$}\label{line:milc_h1_match}
				\State $sv_{v}(m) \gets sv_{v}(M(m))$
				\State $ev_{v}(m) \gets ev_{v}(M(m))$
				\State $st_{v}(m) \gets st_{v}(M(m))$
				\State $et_{v}(m) \gets et_{v}(M(m))$
			\EndFor
			\State $st(v) \gets st(v') + \min_{m \in R_i}{\{ et_{v'}(m) + }$\label{line:milc_h1_update}
			\StatexIndent[3] $dist_{G_M}(ev_{v'}(m), sv_{v}(m)) \}$
			%\StatexIndent[3] $shortest\_path\_length_{G_M}(ev_{v'}(m), sv_{v}(m)) \}$
			\State $v' \gets v$
		\EndFor
	\EndFor
\end{algorithmic}
\end{algorithm}

The first heuristic \mbox{MILC-H1} (Algorithm~\ref{alg:milc_h1}) determines the minimum number of UAVs necessary to achieve $L^{constr}$ for each sensing location $v$ and stores the value in $r_v$ (Line~\ref{line:milc_h1_minrobot}). Function $min\_latency$ returns the minimum-latency that can be achieved with a given number of UAVs $i$ for a path from vertex $v$ to the BS $v_0$. If there is a vertex $v$ with $r_v > |R|$, the problem is infeasible because it is not possible to transport the data with the available number of UAVs within time $L^{constr}$ to the BS (Line~\ref{line:milc_h1_infeasible}). Given the number of UAVs necessary, a TSP tour is split into multiple subtours (Line~\ref{line:milc_h1_maxr} to Line~\ref{line:milc_h1_splittour}). For splitting the tour we use k-SPLITOUR from \cite{Frederickson1976}, which tries to minimize the length of the largest subtour.

The subtours are then traversed by different groups of UAVs (loop in Line~\ref{line:milc_h1_fortours}). For every vertex $v \in V_S$ on a subtour the MLP is calculated with $min\_latency\_path()$, which returns the start and end vertices ($sv_v$ and $ev_v$) and the start and end times ($st_v$ and $et_v$) for every UAV along the MLP (Line~\ref{line:milc_h1_minlatencypath}).

Which UAV should actually move to which start vertex on the MLP for $v \in V_S$ is determined by a matching calculated based on its end vertex $ev_{v'}$ and end time $et_{v'}$ on the MLP for the predecessor $v' \in V_S$ of $v$ on the subtour. The value $st(v)$ determines the time the first UAV on the MLP for $v$ can start to move from $v$ to its end position $ev_v$ and is measured from the beginning of the mission. The start time $st_v$ and end time $et_v$ are relative to the start of the first UAVs ($st_v=0$ for the first UAV). The element $A_{lm}$ of the weight matrix $A$, calculated in the loop starting in Line~\ref{line:milc_h1_calcm}, is the earliest time UAV $m$ can arrive at the potential new starting vertex $sv_v(l)$ after moving from $sv_{v'}(m)$ over $ev_{v'}(m)$ to $sv_{v}(l)$. The matching between UAVs and start vertices minimizes the latest time a UAV can be at its start vertex and is calculated with $minmax\_matching()$ (Line~\ref{line:milc_h1_minmax}). Finally, the mappings are updated according to the matching (Line~\ref{line:milc_h1_match}), and the start value $st(v)$ is calculated based on the latest time all UAV can be at its start vertex (Line~\ref{line:milc_h1_update}). The function $dist_{G_M}(s,d)$ returns the length of the shortest path from $s$ to $d$ in $G_M$. %shortest\_path\_length_{G_M}(s,d)

\begin{figure}[t]
	\centering
	\begin{tabular}{cc}
		\subfloat[]{
			\label{fig:example_min_latency_path}
			\includegraphics[scale=0.39]{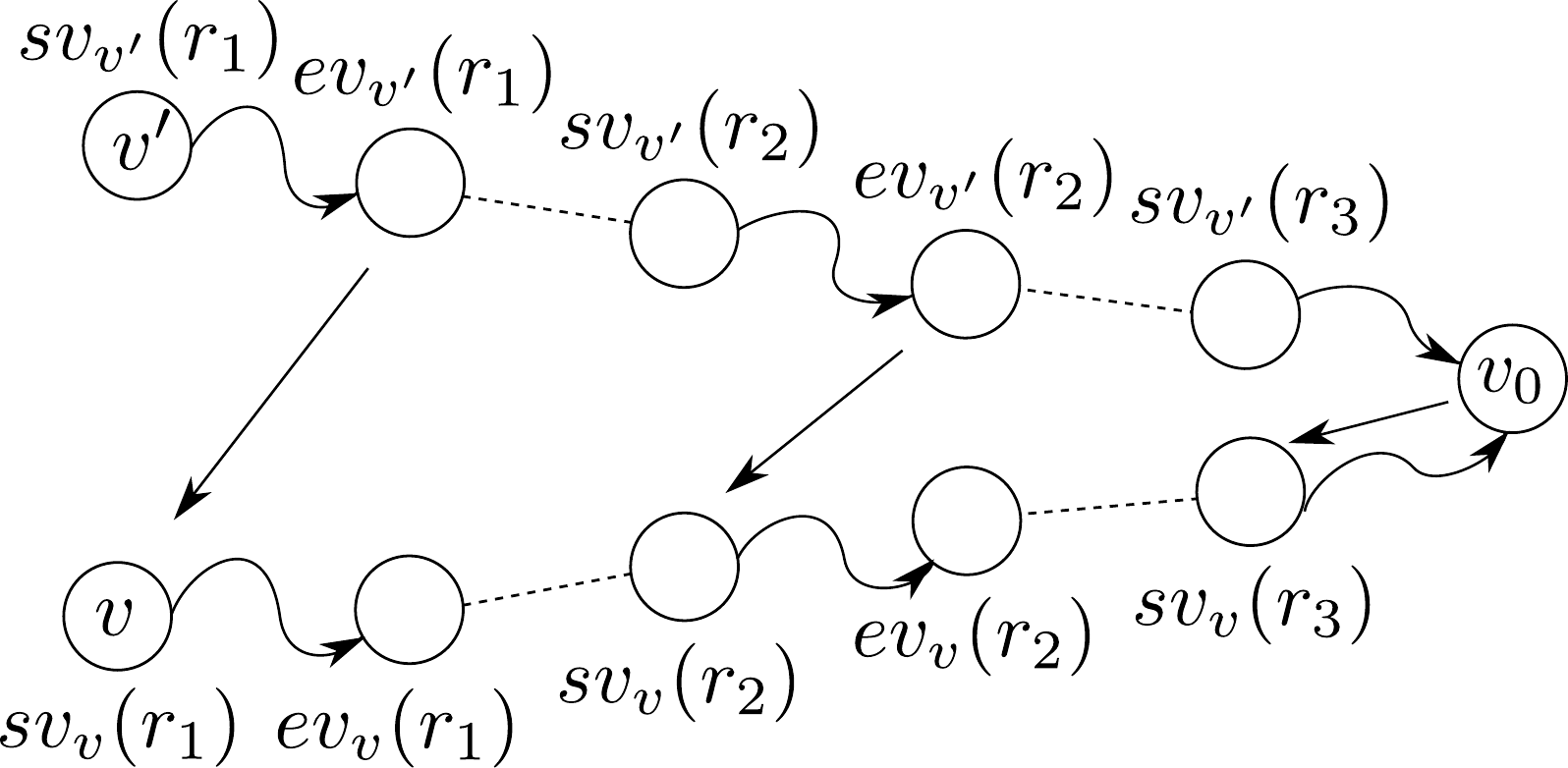}
		}
		\\
		\subfloat[]{
			\label{fig:example_min_latency_timing}
			\includegraphics[scale=0.39]{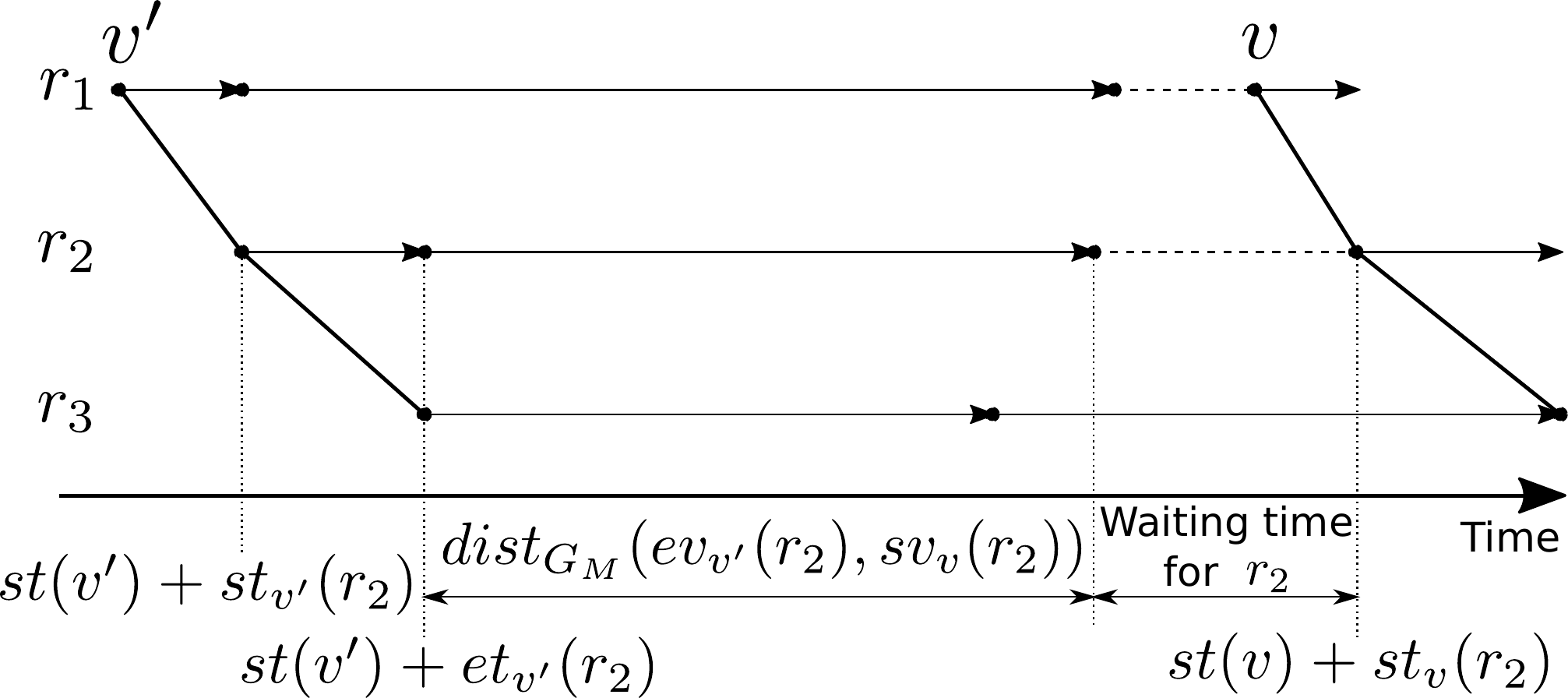}
		}
	\end{tabular}
	\caption{(a) The MLPs for two consecutive sensing locations $v'$ and $v$ on a tour and three UAVs $r_1, r_2$ and $r_3$. The waved lines depict a path in $G_M$ and the dashed lines edges in $G_C$. The straight arrows show the transition from the end vertex $ev_{v'}$ on the MLP for $v'$ to the start vertex $sv_v$ for the MLP of $v$ for each UAV. (b) Timing diagram for the scenario above. Horizontal solid lines denote the movement, horizontal dashed lines denote that a UAV is waiting at a vertex, dots denote the start and end vertex of a UAV on the MLP, and oblique lines depict the MLPs (without transmission times $W^C$, which are 0 in the example). The values on the time axis are labeled only for UAV $r_2$.} %UAVs $r_1$ and $r_2$ have to wait at their start vertices for the MLP for $v$ because the start time for $v$ is determined by the time $r_3$ arrives at its start vertex $sv_v(r_3)$ for the MLP for $v$.
	\label{fig:example_min_latency}
\end{figure}

Figure~\ref{fig:example_min_latency_path} depicts an example situation with two consecutive sensing locations $v'$ and $v$ on a TSP tour (i.e. $v$ gets visited after $v'$) and their MLPs for $3$ UAVs $r_1, r_2$ and $r_3$. For $v'$ UAV $r_1$ starts at $v'=sv_{v'}(r_1)$ and moves to $ev_{v'}(r_1)$ where it transmits the data to $r_2$, which starts at $sv_{v'}(r_2)$ and moves to $ev_{v'}(r_2)$. There, $r_2$ transmits the data to $r_3$. The timing diagram is shown in Figure~\ref{fig:example_min_latency_timing}. The start time $st(v)$ (and therefore the time $r_1$ can start at $v$) is determined by $r_3$, because $r_1$ has to wait such that it does not arrive at $ev_v(r_1)$ before $r_2$ has arrived at $sv_v(r_2)$. UAV $r_2$ in turn has to wait for UAV $r_3$. Note that $r_1$ and $r_2$ can start before $r_3$ reaches $sv_{v}(r_3)$ and $r_2$ will arrive at $ev_{v}(r_2)$ and transmit the data to $r_3$ exactly when the latter one arrives at $sv_{v}(r_3)$. Since the first UAV has to wait at $v \in V_S$, the latency bounds are met because $v$ can be considered as visited right before the first UAV leaves $v$ and the corresponding data will arrive within the bound at the BS.

The second heuristic \mbox{MILC-H2} is similar to the first one. For every sensing location the number of UAVs $r_v$ is calculated such that the latency cannot be decreased on a MLP with additional UAVs (this is different from the loop in Line~\ref{line:milc_h1_loopminrobot} in Algorithm~\ref{alg:milc_h1}). This is equivalent to minimizing the length of the MLP and therefore the latency for each sensing location. If there is a number of UAVs available that is at least a multiple $k$ of $\max_{v\in V_S}{r_v}$, then the tour is split into $k$ subtours. The algorithm then tries for each subtour to visit as many sensing locations as possible with one UAV along the TSP tour before transporting the data with help of the others to the BS such that $L^{constr}$ is not violated. This is different from \mbox{MILC-H1} where UAVs transport the data immediately to the BS after a visit at a sensing location (cf. Line~\ref{line:milc_h1_forpath} in Algorithm~\ref{alg:milc_h1}). Another difference is that in each subtour the same UAV is visiting all sensing locations (i.e. it is not part of the matching, cf. Line~\ref{line:milc_h1_minmax} in Algorithm~\ref{alg:milc_h1}).

The third heuristic \mbox{MILC-H3} is a combination of the other two. For every sensing location the minimum number of UAVs is calculated such that $L^{constr}$ is not violated (similar to \mbox{MILC-H1}). Also, if possible, the tour is split into subtours. The algorithm visits in every subtour as many sensing locations as possible such that $L^{constr}$ is not violated (similar to \mbox{MILC-H2}).

For all heuristics the TSP tour is shortcut if a sensing location has been visited on the MLP of another sensing location already. This is valid because the latency constraint is met for these sensing locations.

The heuristics have $G_C$ with $W^C$ as input, and we therefore assume a constant transmission time on each each edge independent of the amount of transmitted data. This assumption is valid for \mbox{MILC-H1} if the amount of data captured at each sensing location is constant. For the other heuristics the weights $W^C$ could be recalculated after the visit of a sensing location with the consequence that determining whether a feasible solution can be generated, cannot be done before iterating through the vertices of a calculated path, as it is done in Algorithm~\ref{alg:milc_h1} (cf. Line~\ref{line:milc_h1_forpath}).

\section{Simulation results}
\label{sec:eval}

In this section we describe the results from simulation experiments with the aim to assess the performance of our three heuristics in terms of worst idleness $WI$ (maximum idleness over all sensing locations in time steps) in different situations (e.g. number of UAVs, latency bound).

The environment is modeled as rectangular grid of cells of unit size, and time is discretized into time steps. A UAV can move from one cell of the grid to one of the 8 neighboring cells (which determines $G_M$) or stay at the same cell within one time step. The communication range $R^{com}$ (measured in number of cells) determines which cells are within communication range (and therefore determines $G_C$). Unless otherwise stated, we set $W^C$ to zero for all edges. The BS is in the cell at the lower left corner.

A genetic algorithm implementation\footnote{Matlab function tsp\_ga from Joseph Kirk at https://www.mathworks.com/matlabcentral/fileexchange/ 13680-traveling-salesman-problem-genetic-algorithm} is used to generate tours through all sensing locations and the BS, and each experiment is repeated with 10 different tours. To obtain subtours \mbox{k-SPLITOUR} \cite{Frederickson1976} is used.

\begin{figure}[t]
	\centering
	\begin{tikzpicture}
		\begin{axis}[
			height=6cm,
			width=1\columnwidth,
			legend style={at={(1, 1)}, font=\footnotesize, anchor=north east, legend columns=1, legend cell align=left}, 
			ylabel=$WI$,
			xlabel=$L^{constr}$,
			xtick=data,
			x tick label style={rotate=0, anchor=north},
			xtick pos=left,
			ytick pos=left,
			xmin=9, xmax=27,
			ymin=0, ymax=4000,
			axis lines*=left
			]

			%W=400 (starting at (6,6)), tour generated by tsp_ga(pop=1500, iter=15000), U=6, Rcom=4
			%H1
			\addplot[mark options={scale=0.6,solid}]%, error bars/y dir=both, error bars/y explicit]
			table[x index=0, y index=1, y error index=2] {data/shortcut/midc_ddata.dat}
			[yshift=-8pt]
					%pos are not linear on the x-axis but relative to the plot line length
					node[pos=0   ]{$1$}
					node[pos=0.14]{$1$}
					node[pos=0.27]{$1$}
					node[pos=0.41]{$1$}
					node[pos=0.54]{$2$}
					node[pos=0.63]{$3$}
					node[pos=0.8 ]{$3$}
					node[pos=0.98]{$6$}
					node[pos=1   ]{$6$}
				;

			%coordinates { %(Ddata, WI), R
				%(10, 3772)	+-( 21,  21)	%1
				%(12, 4355)	+-( 11,  11)	%1
				%(14, 4895)	+-(  4,   4)	%1
				%(16, 5409)	+-(  5,   5)	%1
				%(18, 5560)	+-(336, 336)	%2
				%(20, 4101)	+-(291, 291)	%3
				%(22, 4309)	+-(376, 376)	%3
				%(24, 2455)	+-(144, 144)	%6
				%(26, 2455)	+-(144, 144)	%6
				%}
				%[yshift=-9pt]
					%pos are not linear on the x-axis but relative to the plot line length
					%node[pos=0    ]{$1$}
					%node[pos=0.11 ]{$1$}
					%node[pos=0.215]{$1$}
					%node[pos=0.307]{$1$}
					%node[pos=0.34 ]{$2$}
					%node[pos=0.61 ]{$3$}
					%node[pos=0.66 ]{$3$}
					%node[pos=0.98 ]{$6$}
					%node[pos=1    ]{$6$}
				%;
			
			%H2
			\addplot[style=densely dashed, mark options={scale=0.6,solid}]%, error bars/y dir=both, error bars/y explicit]
			%table[x index=0, y index=7, y error index=8] {data/shortcut/midc_ddata.dat};
			coordinates { %(Ddata, WI)
				(10, 1328)	+-(29, 29)	%1
				(12, 1139)	+-(37, 37)	%1
				(14,  935)	+-(41, 41)	%1
				(16,  745)	+-(55, 55)	%1
				(18,  583)	+-(62, 62)	%1
				(20,  533)	+-(65, 65)	%1
				(22,  460)	+-(42, 42)	%1
				(24,  437)	+-(22, 22)	%1
				(26,  412)	+-( 8,  8)	%1
				};

			%H3
			\addplot[style=loosely dashed, mark options={scale=0.6,solid}]%, error bars/y dir=both, error bars/y explicit]
			table[x index=0, y index=13, y error index=14] {data/shortcut/midc_ddata.dat};
			%coordinates { %(Ddata, WI)
				%(10, 2325)	+-( 39,  39)	%1
				%(12, 2286)	+-( 28,  28)	%1
				%(14, 2314)	+-( 67,  67)	%1
				%(16, 2732)	+-( 73,  73)	%1
				%(18, 2772)	+-(618, 618)	%2
				%(20, 1943)	+-(286, 286)	%3
				%(22, 1449)	+-(218, 218)	%3
				%(24,  553)	+-( 43,  43)	%6
				%(26,  418)	+-( 34,  34)	%6
				%};

			\legend{MILC-H1\\MILC-H2\\MILC-H3\\}
			
		\end{axis}
	\end{tikzpicture}
	
	\caption{$WI$ for varying $L^{constr}$ for the \mbox{MILC} heuristics with $n=6$ and $R^{com}=4$. The numbers below the plots show the numbers of subtours for \mbox{MILC-H1} and \mbox{MILC-H3} (for \mbox{MILC-H2} it is always $1$).}
	\label{fig:res_milc_ddata_wi}
\end{figure}
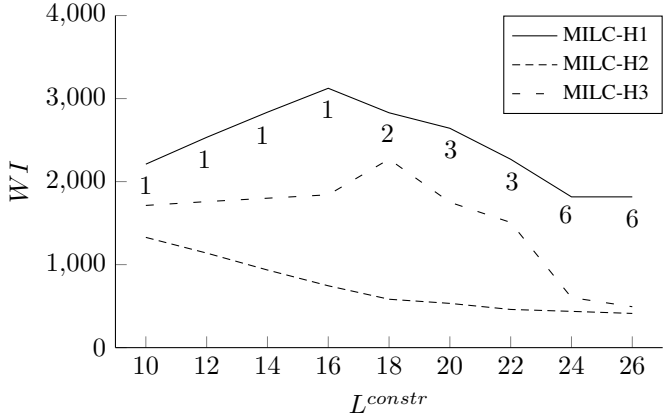

Figure~\ref{fig:res_milc_ddata_wi} shows a comparison of $WI$ between the heuristics \mbox{MILC-H1}, \mbox{MILC-H2}, and \mbox{MILC-H3} for varying latency constraint $L^{constr}$ on a grid with $20 \times 20$ sensing locations. The number of UAVs $n$ is $6$ and the communication range $R^{com}=4$. The counter intuitive behavior of \mbox{MILC-H1} and \mbox{MILC-H3}, which show an increasing $WI$ with increasing $L^{constr}$, results from the fact that the algorithm minimize the number of UAVs necessary to meet the latency constraints for each sensing location. This results in longer durations for the transportation of the data to the BS. The drops in $WI$ happen when the number of subtours increase due to a splitting of the original tour. \mbox{MILC-H2} shows the expected behavior of an decreasing $WI$ with increasing $L^{constr}$ because with a larger latency bound more sensing locations can be visited consecutively before transporting the data to the BS. \mbox{MILC-H2} is not able to split the tour with the available number of UAVs since in all cases the latency could be further decreased with an increasing number of UAVs. For \mbox{MILC-H3} the number of subtours is the same as for \mbox{MILC-H2}. In contrast to \mbox{MILC-H1}, \mbox{MILC-H3} can benefit from an increasing latency bound since more sensing locations can be visited without data transportation to the BS.

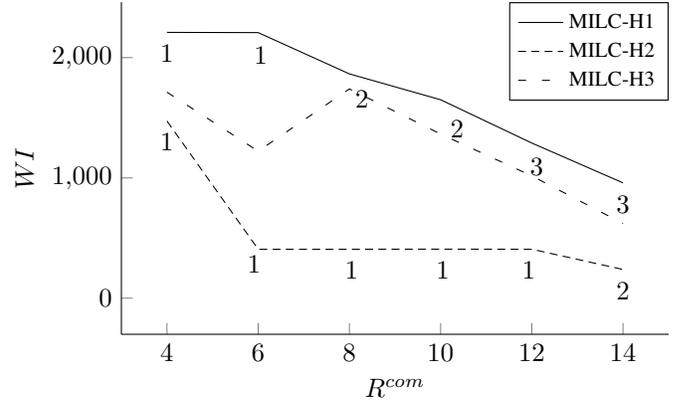
\begin{figure}[t]
	\centering
	\begin{tikzpicture}
		\begin{axis}[
			height=6cm,
			width=1\columnwidth,
			legend style={at={(1, 1)}, font=\footnotesize, anchor=north east, legend columns=1, legend cell align=left}, 
			ylabel=$WI$,
			xlabel=$R^{com}$,
			xtick=data,
			x tick label style={rotate=0, anchor=north},
			xtick pos=left,
			ytick pos=left,
			xmin=3, xmax=15,
			ymin=-300,
			axis lines*=left
			]

			%W=400 (starting at (6,6)), tour generated by tsp_ga(pop=1500, iter=15000), U=6, Ddata=10
			%H1
			\addplot[mark options={scale=0.6,solid}]%, error bars/y dir=both, error bars/y explicit]
			table[x index=0, y index=1, y error index=2] {data/shortcut/midc_rcom.dat}
			[yshift=-8pt]
					%pos are not linear on the x-axis but relative to the plot line length
					node[pos=0   ]{$1$}
					node[pos=0.01]{$1$}
					node[pos=0.3 ]{$2$}
					node[pos=0.5 ]{$2$}
					node[pos=0.75]{$3$}
					node[pos=1   ]{$3$}
			;
			%coordinates { %(Rcom, WI), R
				%( 4, 3772)	+-( 21,  21)%1
				%( 6, 3327)	+-(  9,   9)%1
				%( 8, 2904)	+-( 85,  85)%2
				%(10, 2524)	+-(100, 100)%2
				%(12, 1585)	+-( 75,  75)%3
				%(14, 1465)	+-(137, 137)%3
				%}
				%[yshift=-8pt]
					%pos are not linear on the x-axis but relative to the plot line length
					%node[pos=0    ]{$1$}
					%node[pos=0.19 ]{$1$}
					%node[pos=0.375]{$2$}
					%node[pos=0.54 ]{$2$}
					%node[pos=0.945]{$3$}
					%node[pos=1    ]{$3$}
				%;
			
			%H2
			\addplot[style=densely dashed, mark options={scale=0.6,solid}]%, error bars/y dir=both, error bars/y explicit]
			table[x index=0, y index=7, y error index=8] {data/shortcut/midc_rcom.dat}
			[yshift=-8pt]
					%pos are not linear on the x-axis but relative to the plot line length
					node[pos=0     ]{$1$}
					node[pos=0.82  ]{$1$}
					node[pos=0.861 ]{$1$}
					node[pos=0.8628]{$1$}
					node[pos=0.8645]{$1$}
					node[pos=1     ]{$2$}
			;
			%coordinates { %(Rcom, WI), R
				%( 4, 1328)	+-(29, 29)%1
				%( 6,  407)	+-( 3,  3)%1
				%( 8,  407)	+-( 3,  3)%1
				%(10,  408)	+-( 2,  2)%1
				%(12,  407)	+-( 3,  3)%1
				%(14,  239)	+-(13, 13)%2
				%}
				%[yshift=-8pt]
					%pos are not linear on the x-axis but relative to the plot line length
					%node[pos=0      ]{$1$}
					%node[pos=0.82   ]{$1$}
					%node[pos=0.84255]{$1$}
					%node[pos=0.8446 ]{$1$}
					%node[pos=0.85   ]{$1$}
					%node[pos=1      ]{$2$}
				%;

			%H3
			\addplot[style=loosely dashed, mark options={scale=0.6,solid}]%, error bars/y dir=both, error bars/y explicit]
			table[x index=0, y index=13, y error index=14] {data/shortcut/midc_rcom.dat}
			%[yshift=-8pt]
					%pos are not linear on the x-axis but relative to the plot line length
					%node[pos=0    ]{$1$}
					%node[pos=0.255]{$1$}
					%node[pos=0.44 ]{$2$}
					%node[pos=0.665]{$2$}
					%node[pos=0.865]{$3$}
					%node[pos=1    ]{$3$}
			;
			%coordinates { %(Rcom, WI), R
				%( 4, 2325)	+-( 39,  39)%1
				%( 6, 1654)	+-( 56,  56)%1
				%( 8, 2143)	+-(250, 250)%2
				%(10, 1542)	+-(334, 334)%2
				%(12, 1019)	+-(150, 150)%3
				%(14,  664)	+-( 49,  49)%3
				%}
				%[yshift=-8pt]
					%pos are not linear on the x-axis but relative to the plot line length
					%node[pos=0    ]{$1$}
					%node[pos=0.255]{$1$}
					%node[pos=0.44 ]{$2$}
					%node[pos=0.665]{$2$}
					%node[pos=0.865]{$3$}
					%node[pos=1    ]{$3$}
				%;

			\legend{MILC-H1\\MILC-H2\\MILC-H3\\}
			
		\end{axis}
	\end{tikzpicture}
	\caption{$WI$ of MILC heuristics for varying $R^{com}$, $n=6$, $L^{constr}=10$. The numbers below the plots show the numbers of subtours (same for \mbox{MILC-H1} and \mbox{MILC-H3}).}
	\label{fig:res_milc_rcom_wi}
\end{figure}

Figure~\ref{fig:res_milc_rcom_wi} shows a comparison between the three MILC heuristics for a communication range $R^{com}$ from $4$ to $14$ cells with $n=6$ and $L^{constr}=10$. The tight constraints force the UAVs to frequently transport the data to the BS. Compared to \mbox{MILC-H3}, \mbox{MILC-H2} benefits from minimizing the latency by maximizing the number of UAVs for the MLP, which leads to a faster progress along the TSP tour.

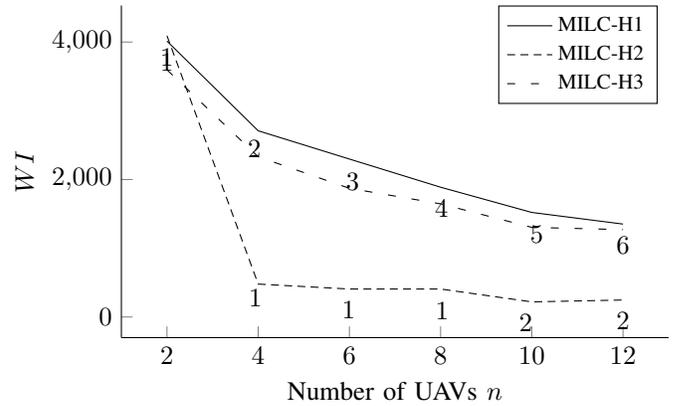
\begin{figure}[t]
	\centering
	\begin{tikzpicture}
		\begin{axis}[
			height=6cm,
			width=1\columnwidth,
			xmin=1, xmax=13,
			ymin=-300,
			axis y line*=left,
			axis x line*=bottom,
			xtick=data,
			xlabel=Number of UAVs $n$,
			ylabel=$WI$,
			legend style={at={(0.98, 1)}, font=\footnotesize, anchor=north east, legend columns=1, legend cell align=left}
			]

			%W=400 (starting at (6,6)), tour generated by tsp_ga(pop=1500, iter=15000), Rcom=4, dataex_full=false
			%H1
			\addplot[mark options={scale=0.6,solid}]%, error bars/y dir=both, error bars/y explicit]
			table[x index=0, y index=1, y error index=2] {data/shortcut/midc_u.dat}
			[yshift=-8pt]
					%pos are not linear on the x-axis but relative to the plot line length
					node[pos=0   ]{$1$}
					node[pos=0.47]{$2$}
					node[pos=0.65]{$3$}
					node[pos=0.8 ]{$4$}
					node[pos=0.94]{$5$}
					node[pos=1   ]{$6$}
			;
			%coordinates { %(U, WI)
				%(2, 6190)
				%(4, 3518)
				%(6, 2386)
				%(8, 1774)
				%(10, 1483)
				%(12, 1220)
				%};

			%H2
			\addplot[style=densely dashed, mark options={scale=0.6,solid}]%, error bars/y dir=both, error bars/y explicit]
			table[x index=0, y index=7, y error index=8] {data/shortcut/midc_u.dat}
			[yshift=-8pt]
					%pos are not linear on the x-axis but relative to the plot line length
					node[pos=0     ]{$1$}
					node[pos=0.9   ]{$1$}
					node[pos=0.9445]{$1$}
					node[pos=0.946 ]{$1$}
					node[pos=0.99  ]{$2$}
					node[pos=1     ]{$2$}
			;
			%coordinates { %(U, WI)
				%( 2, 407)	+-(3, 3)
				%( 4, 407)	+-(3, 3)
				%( 6, 407)	+-(3, 3)
				%( 8, 407)	+-(3, 3)
				%(10, 407)	+-(3, 3)
				%(12, 407)	+-(3, 3)
				%};

			%H3
			\addplot[style=loosely dashed, mark options={scale=0.6,solid}]%, error bars/y dir=both, error bars/y explicit]
			table[x index=0, y index=13, y error index=14] {data/shortcut/midc_u.dat}
			%[yshift=-8pt]
					%pos are not linear on the x-axis but relative to the plot line length
					%node[pos=0     ]{$1$}
					%node[pos=0.82  ]{$2$}
					%node[pos=0.861 ]{$3$}
					%node[pos=0.8628]{$4$}
					%node[pos=0.8645]{$5$}
					%node[pos=1     ]{$6$}
			;
			%coordinates { %(U, WI)
				%( 2, 242)	+-(15, 15)
				%( 4, 155)	+-( 6,  6)
				%( 6, 116)	+-( 8,  8)
				%( 8, 104)	+-(14, 14)
				%(10,  91)	+-(16, 16)
				%(12,  86)	+-(12, 12)
				%};

			\legend{MILC-H1\\MILC-H2\\MILC-H3\\}
			
		\end{axis}
	\end{tikzpicture}
	
	\caption{$WI$ of the heuristics for varying number of UAVs $n$, $L^{constr}=14$, $R^{com}=8$. The numbers below the plots show the numbers of subtours (same for \mbox{MILC-H1} and \mbox{MILC-H3}).}
	\label{fig:res_milc_u_wi}
\end{figure}

The results for a varying number of UAVs is shown in Figure~\ref{fig:res_milc_u_wi} with $L^{constr}=14$ and $R^{com}=8$ such that at least 2 UAVs are necessary to transport the data within $L^{constr}$ to the BS. The large drop for \mbox{MILC-H2} happens because $R^{com}$ is large enough that one UAV can move nearly freely on the tour without waiting for the other UAVs to transport the data.

\begin{figure}[t]
	\centering
	\begin{tikzpicture}
		\begin{axis}[
			height=6cm,
			width=1\columnwidth,
			xmin=-1, xmax=6,
			ymin=-300,
			axis y line*=left,
			axis x line*=bottom,
			xtick=data,
			xlabel=$W^C_{ij}$,
			ylabel=$WI$,
			legend style={at={(0.1, 1)}, font=\footnotesize, anchor=north west, legend columns=1, legend cell align=left}
			]

			%W=400 (starting at (6,6)), tour generated by tsp_ga(pop=1500, iter=15000), Rcom=4, dataex_full=false
			%H1
			\addplot[mark options={scale=0.6,solid}]%, error bars/y dir=both, error bars/y explicit]
			table[x index=0, y index=1, y error index=2] {data/shortcut/milc_ltrans.dat}
			[yshift=-8pt]
					%pos are not linear on the x-axis but relative to the plot line length
					node[pos=0   ]{$6$}%0
					node[pos=0.06]{$6$}%1
					node[pos=0.12]{$6$}%2
					node[pos=0.175]{$6$}%3
					node[pos=0.26]{$6$}%4
					node[pos=1  ]{$3$}%5
			;

			%H2
			\addplot[style=densely dashed, mark options={scale=0.6,solid}]%, error bars/y dir=both, error bars/y explicit]
			table[x index=0, y index=7, y error index=8] {data/shortcut/milc_ltrans.dat}
			%[yshift=-8pt]
					%%pos are not linear on the x-axis but relative to the plot line length
					%node[pos=0     ]{$1$}
					%node[pos=0.9   ]{$1$}
					%node[pos=0.9445]{$1$}
					%node[pos=0.946 ]{$1$}
					%node[pos=0.99  ]{$2$}
					%node[pos=1     ]{$2$}
			%
			;

			%H3
			\addplot[style=loosely dashed, mark options={scale=0.6,solid}]%, error bars/y dir=both, error bars/y explicit]
			table[x index=0, y index=13, y error index=14] {data/shortcut/milc_ltrans.dat}
			%[yshift=-8pt]
					%pos are not linear on the x-axis but relative to the plot line length
					%node[pos=0     ]{$1$}
					%node[pos=0.82  ]{$2$}
					%node[pos=0.861 ]{$3$}
					%node[pos=0.8628]{$4$}
					%node[pos=0.8645]{$5$}
					%node[pos=1     ]{$6$}
			;

			\legend{MILC-H1\\MILC-H2\\MILC-H3\\}
			
		\end{axis}
	\end{tikzpicture}
	
	\caption{$WI$ of the heuristics for varying transmission times $W^C_{ij}$, $n=6$, $L^{constr}=24$, $R^{com}=8$. The numbers below the plots show the numbers of subtours for \mbox{MILC-H1} and \mbox{MILC-H3} (for \mbox{MILC-H2} it is always $1$).}
	\label{fig:res_milc_ltrans_wi}
\end{figure}
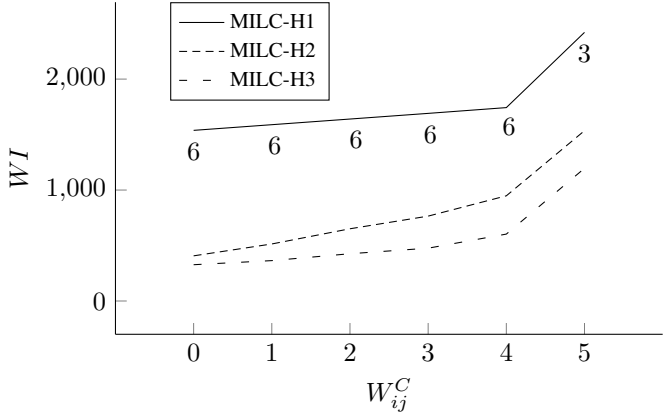

Finally, we compare the performance for different transmission times $W^C_{ij}$ (the same value for all edges of $E_C$). The results for $n=6$, $L^{constr}=24$ (such that one and two UAVs are necessary to meet the constraint), and $R^{com}=8$ is shown in Figure~\ref{fig:res_milc_ltrans_wi}. The effect of an increasing transmission time is larger for \mbox{MILC-H2} because the algorithm tries to use as many UAVs as possible for the MLP and the transmission time sums up along the MLP. In contrast to this, \mbox{MILC-H1} and \mbox{MILC-H3} use one UAV which is sufficient to transport the data within the bound for $W^C_{ij} \leq 4$. For this relative high value (compared to the previous simulations) of $L^{constr}$, \mbox{MILC-H3} performs best because it is possible to visit several sensing locations with one UAV before it has to return to the BS, and therefore the tour can be split into multiple subtours.

To summarize our simulations, the simplest heuristic MILC-H1 performs worst in all experiments and serve as baseline, whereas \mbox{MILC-H2} outperforms the other heuristics with the expected behavior of a decreasing $WI$ with increasing $L^{constr}$, $R^{com}$ and $n$ and therefore exhibits the behavior of individual and cooperative data transport. In all experiments we used low values for $L^{constr}$ to see the effect of the constraint on the algorithms. As $L^{constr}$ increases, the worst idleness $WI$ for \mbox{MILC-H2} and \mbox{MILC-H3} approaches the optimal value given the TSP tour which results in a solution where each UAV transports the data to the BS individually.

\section{Conclusion}
\label{sec:conclusion}

We presented a multi-UAV surveillance problem with cooperative data transport. The aim is to minimize the idleness of sensing locations while constraining the latency to a predefined bound. This enforces the UAVs to transport the data cooperatively to the BS in a store-and-forward fashion. We presented different heuristics that are based on MLPs. MLPs guarantee that the data is delivered in due time within a predefined latency bound, which is a typical requirement in disaster response scenarios. We evaluated the performance in simulation experiments which show that the baseline heuristic \mbox{MILC-H1} performs worst, whereas \mbox{MILC-H2} outperforms the other heuristics with the expected behavior of a decreasing worst idleness with increasing values of the latency bound, the communication range, and the number of robots when the transmission time is zero.

The algorithms rely on TSP tours through all sensing locations which have been generated with traditional algorithms that try to minimize the length of the tour. An open issue is the generation of such tours that support the joint minimization of idleness and latency. Another possible improvement for the heuristics is a more sophisticated scheduling of UAVs to minimize the number of idle UAVs (that do neither sensing nor transporting data) at each time step. The team of UAVs is divided into a fixed partition of teams and only one UAV in a team does sensing while the other UAVs transport the data to the BS although not all of them might be necessary to meet the latency constraint. Scalability and robustness is limited by the fact that a centralized entity has to generate the solution before the mission starts.

%%%%%%%%%%%%%%%%%%%%%%%%%%%%%%%%%%%%%%%%%%%%%%%%%%%%%%%%%%%%%%%%%%%%%%%%%%%%%%%%

\bibliographystyle{abbrv}
\bibliography{IEEEabrv,references}

\end{document}